\documentclass[final]{cvpr}

\usepackage{times}
\usepackage{epsfig}
\usepackage{graphicx}
\usepackage{amsmath}
\usepackage{amssymb}

\usepackage[pagebackref=true,breaklinks=true,colorlinks,bookmarks=false]{hyperref}

\def\cvprPaperID{61} 
\def\confYear{CVPR 2021}

\begin{document}

\title{Three-stream network for enriched Action Recognition}

\author{Ivaxi Sheth\\
Imagination Technologies\\
{\tt\small ims116@ic.ac.uk}
}

\maketitle

\begin{abstract}
Understanding accurate information on human’s behaviours is one of the most important tasks in machine intelligence. Human Activity Recognition that aims to understand human activities from a video is a challenging task due to various problems including background, camera motion and dataset variations. This paper proposes two CNN based architectures with three streams which allow the model to exploit the dataset under different settings. The three pathways are differentiated in frame rates. The single pathway, operates at a single frame rate captures spatial information, the slow pathway operates at low frame rates captures the spatial information and the fast pathway operates at high frame rates that capture fine temporal information. Post CNN encoders, we add bidirectional LSTM and attention heads respectively to capture the context and temporal features. By experimenting with various algorithms on UCF-101, Kinetics-600 and AVA dataset, we observe that the proposed models achieve state-of-art performance for human action recognition task. \end{abstract}

\section{Introduction}

As imaging systems become ubiquitous, the ability to recognize human actions is becoming increasingly important.
\textbf{Activity Recognition} is an elemental task in the computer vision  that realises human actions. These actions are detected post complete action execution in a video. Through this approach, action and its purpose can be identified. Applications of AR are becoming highly recognisable within surveillance, video retrieval, human robot interactions and self driving vehicles. 

From a group of videos $S$ and its corresponding action labels $L$, each video $V \in S$ contains one or more actions $l_V$. Thus the aim of activity recognition problem is to predict labels $l_V$ based on video understanding $V$. In case of action recognition, the video is generally segmented to contain only one execution of a human action. In more general cases, the video may contain multiple actions and the goal of action detection is not just to recognize the actions being performed in the video but also determine the spatio-temporal boundaries of them. 

One of the reasons why the transition from 2D images to 3D videos hasn't been smooth is lack of temporal understanding. This paper proposes network that enhances not only temporal features but also spatial features. The spatial dimensions X and Y in 2D images are considered with equal importance.\cite{feichtenhofer2019slowfast} paper introduces that time dimension, T in a video is not as important as the spatial dimensions, X and Y. Hence our paper exploits this property to perform action recognition from videos.

\begin{figure*}[h]
\begin{center}
  \includegraphics[width=1.0\linewidth]{ 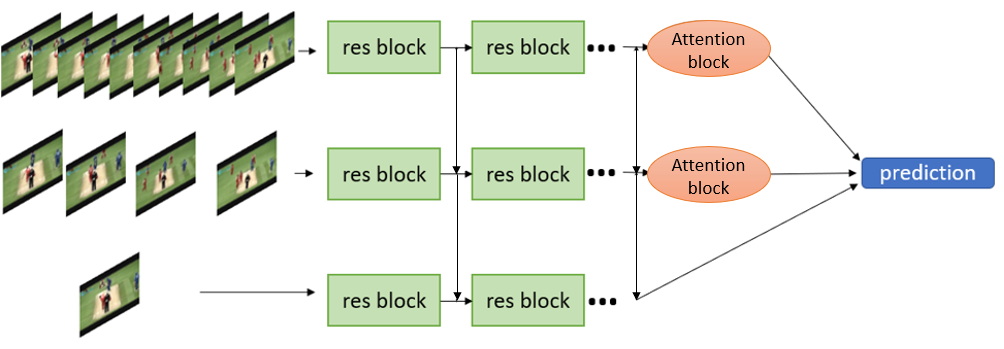}
\end{center}
   \caption{An overview of our three stream network with attention head. The network architecture for our baseline i.e. three stream network with bi-LSTM follows similar architecture with bidirectional LSTM block instead of Attention block}
\label{fig:short}
\end{figure*}

Our method proposes a three lane network where each pathway is differentiated in frame rates. The single pathway, operates at single frame rate captures spatial information, the slow pathway operates at low frame rates captures the spatial information and the fast pathway operates at high frame rates that captures fine temporal information. Inspired from textual tasks, we propose two networks one with bi-directional LSTM head (we call it baseline) and other with attention mechanism to achieve state of art accuracy. We test our models on multiple datasets, UCF-101 \cite{soomro2012ucf101}, Kinetics-600 \cite{carreira2017quo} and  Atomic Visual Actions (AVA) \cite{gu2018ava}. AVA is a particularly difficult dataset as it requires detecting multiple people in videos semi-densely in time, and recognizing multiple basic actions. With three stream network and bi-directional LSTM head we achieve state of art accuracy for UCF-101 and Kinetics dataset while the three stream network with attention head outperforms all methods for all of 3 datasets. 

\section{Related Works}
Traditionally, for video action recognition, human-created features, like Histogram of Oriented Gradients (HOG)\cite{hog} and Histogram of Optical Flow (HOF)\cite{hof} were used by the research community. These features can be either sparsely\cite{hof} and densely\cite{dense} sampled. These early methods consider independent interest points across frames.

In recent times, deep learning has brought remarkable improvements to image understanding\cite{krizhevsky2012imagenet}. The high performance from image classification networks has propelled these networks to inspire networks for video understanding \cite{karpathy2014large} with minimal modifications. One of the methods, extracts features independently from each frame of the video, applies image classification CNN and finally pools predictions across all the frames of the video. The drawback of this method is that it ignores the temporal structure example: models can't conceivably recognize opening an entryway from shutting an entryway. Adding a recurrent layer to CNN can encode state although unrolling RNNs.

An interesting and practical model was proposed by\cite{2str}. Here a 2D ConvNet was used on short temporal snapshots of videos. The output from single RGB frame and stack of 10 optical flow frames were averaged, after passing them through 2D ConvNet. This model appeared to get extremely superior on existing datasets and computationally efficient as well.

3D Convolution Networks are important to video modeling, and are similar standard convolutional networks, but with spatio-temporal filters. 3D CNNs have an key attribute: ability to generate hierarchical representations of the input video data. Temporal Segment Network \cite{tsn} is based on the idea of long-range temporal structure modelling. It was successful at tackling limited information to temporal context, as the other networks operated on unit frames or a single stack of multiple frames. Instead of working on short snippets, TSN work on short snippets sparsely sampled from the entire video. 

\subsection{Two stream networks}

Two stream networks were introduced by \cite{2str} for action recognition in videos. They used two CNNs for each spatial and temporal information in videos. In this architecture, RGB images are extracted from frames of video and optical flow from consecutive frames are fed into different streams. This work was then expanded to give inflated two stream 3D network.
This work \cite{lan2017deep} proposed to use neural networks along with shallow local features. \cite{2str} proposed a spatiotemporal architecture and explored various layer fusion schemes. They claimed that fusing last convolution layers spatially increases accuracy. \cite{tsn} introduced a neural network performs pixel level action recognition and segmentation by adding a two-stream network along with temporal aggregation. \cite{feichtenhofer2019slowfast} proposes a two stream network where CNN on each stream works on different frame rates.

\subsection{Temporal Understanding}
2D image models have also been extended to videos as seen in \cite{carreira2017quo}.
Long-term filtering and pooling using temporal strides \cite{varol2017long} were then used. Video compression methods were also used \cite{wu2018compressed} for a faster network. Extending on 3D CNNs, some works have separated 2D spatial and 1D temporal filters \cite{tran2018closer}. Eventually 2D image classification networks were inflated to 3D extended to action recognition. 
Our work proposes to perform temporal filtering at different rates along with adding extras blocks to understand the temporal structure.

\subsection{Bidirectional LSTM}
Since RNNs can encode long-term sequence information in a data, they have recently been added into action recognition tasks. Bidirectional LSTMs add a hidden layer that allows information to flow in background direction. Therefore output layer receives information from past and future states simultaneously. Thus Bi-LSTM \cite{graves2013hybrid} increases the amount of input data that can be ingested by network by the network. CNNs have restrictions on the input information adaptability, as they require fixed input data.  A normal RNN also has restriction as upcoming input data can not be attained from the present state. On the contrast, Bi-LSTM don't require the fixed input data and their future input information can be accessed from the current state. 

In literature, LSTMs have been used seldom. \cite{yue2015beyond} connects LSTM cells to CNN feature maps for action recognition. \cite{ullah2017action} proposes to use bi-LSTM along with a convolutional encoder.

\subsection{Attention models}
The concept of attention was introduced by Bahdanau, Cho, and Bengio (2014)
for the objective of machine translation.
Attention mechanism is based on concept that the nwural network learns how admissible are some feature maps with regarding to output state.
These values of importance are specified as weights of attention
and are generally calculated at the same time as other model
parameters trained for a specific goal.

Attention was used in first person action recognition by having a joint learning of gaze and actions \cite{li2018eye}, by using object-centric for ego centric activity recognition \cite{sudhakaran2018attention} and by event modulated attention \cite{hu2018squeeze}. 
Attention was used in to extract spatial information by generating spatial masks by traning on video labels\cite{zhang2018image}. Temporal attention was used for action recognition by detecting change in gaze \cite{shen2018reinforced}.

\section{Methodology}

Since prior works weren't able to extract temporal information effectively, we explore the network under multiple temporal settings namely bidirectional LSTM and Attention. For better spatial understanding we have a combined three stream CNN based network as encoder. We have our baseline (three stream network with bi-LSTM) perform well over UCF-101 and Kinetics dataset. To further improve performance, we use with attention mechanism that gives state-of-art performance over all three datasets.

\subsection{Three stream}
Section 2 discusses works relating to two stream networks. These (and current) methods cannot unfortunately show decent performance on a difficult dataset like AVA. Hence we suggest a novel architecture with three streams.

\subsubsection{Single pathway}
The main aim of single pathway is extract spatial features. The network can have any 2D architecture. We see \cite{karpathy2014large} that single 2D net can extract a great deal of spatial feature. This pathways is essentially equivalent to having 3D network with temporal stride of $\theta_1$, where $\theta_1$ is the equivalent to the speed at which the video is processed. Hence for 30fps video, the stride is 30, therefore ingesting only first frame. This is a very light weight stream with much lower MACs as compared to other streams. 

\subsubsection{Slow pathway}
The main aim is to extract spatio-temporal features. Here the temporal stride, $\theta_2$ is smaller than $\theta_1$. After experimenting as mentioned in section 7, the best value for $\theta_2$ is 16, also as mentioned in paper \cite{feichtenhofer2019slowfast}. This means that approx. 2 frames in a 30fps input video are processed.

\subsubsection{Fast pathway}
Similar to slow pathway, this is a 3d CNN but with much smaller temporal stride to particularly encode for more temporal info. To understand finer representations, we select alternate frames, making stride 2. Hence $\theta_3$ < $\theta_2$ < $\theta_1$.
one of the key features of three stream network is different values of $\theta$ that works on different temporal speeds, and thus drives the expertise of the two subnets instantiating the two pathways. Our Fast pathway also distinguishes with existing models in that it can use significantly lower channel capacity. The low channel capacity can also be interpreted as a weaker ability of representing spatial semantics.

\subsection{Bi-directional LSTM}

In bi-directional LSTM \cite{ullah2017action}, the output at time $t$ is not only dependent on the previous frames in the sequence, but also on the upcoming frames. In our work, we use multiple LSTM layers, so our scheme has two LSTM layers for both forward pass and backward pass. Figure 2 shows the overall concept of bidirectional LSTM used in the proposed method. The input data is fed to the bidirectional RNN, and the hidden states of forward pass and backward pass are combined in the output layer. Both forward and backward consist of two LSTM cells, making our model a deeper. The proposed  method outperforms other state-of-the-art methods due to its  mechanism of computing the output. The output of a frame  at time $t$ is calculated from the previous frame at time $t-1$ and the upcoming frame at time $t + 1$ because layers are 
performing processing in both directions. 

\begin{figure}[h]
\begin{center}
  \includegraphics[width=1.0\linewidth]{ 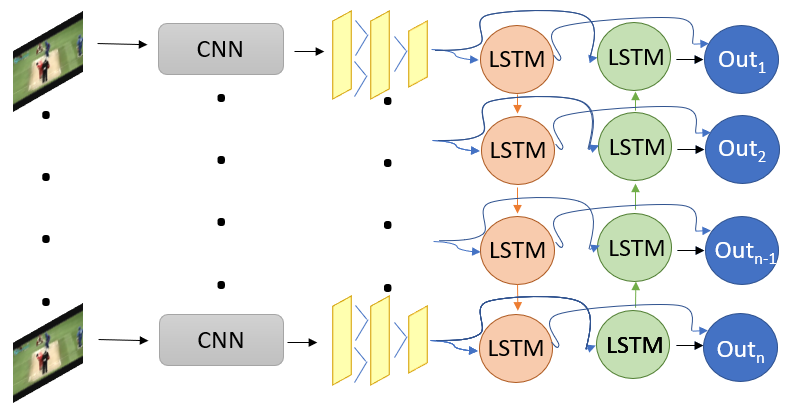}
\end{center}
  \caption{Our bi-directional LSTM head }
\label{fig:long}
\label{fig:onecol}
\end{figure}

\subsection{Attention head}

We use the self mechanism explained in \cite{plizzari2020spatial} for our attention head.
Our self-attention head computes correlation between arbitrary positions of a sequence input. An attention function consists of a query $A_Q$, keys $A_K$ and values $A_V$. The query and keys have same vector dimension $d_k$, and values and outputs have same size, $d_v$. The output is computed as a weighted sum of the values and its weight computed by scaled dot-product of query and keys. The attention function is defined as
\begin{center}
    $Attention(Q,K,V) = softmax(\frac{QK^T}{\sqrt{d_k}})$
\end{center}
$\sqrt{d_k}$ is a scaling factor.

The equation computes scaled dot-product attention and the network computes the attention multiple times in parallel (multi-head) to extract different correlation information. The multi-head attention outputs are concatenated and transformed to the same vector
dimension the input sequence. A residual connection is
adopted to take the input and output of the multi-head self attention layer and a layer normalization is applied to the
summed output. A fully-connected feed-forward network
with a residual connection is applied to the normalized self-attention output. 




\section{Network Architecture}

A major advantage of this skeleton is that one can use any network architecture for each of the streams. We use inflated ResNet \cite{resnet} as
the 3D backbone network, for its promising performance on
various datasets \cite{res3d}. We have studied performance difference for various other network backbones to three stream in Ablation study. Meanwhile, original ResNet-52 serves
as our 3D backbone. We use the output features of res2, res3,
res4, res5 to build our network, where they are spatially downsampled respectively.

\begin{figure}[h]
\begin{center}
  \includegraphics[width=1\linewidth]{ 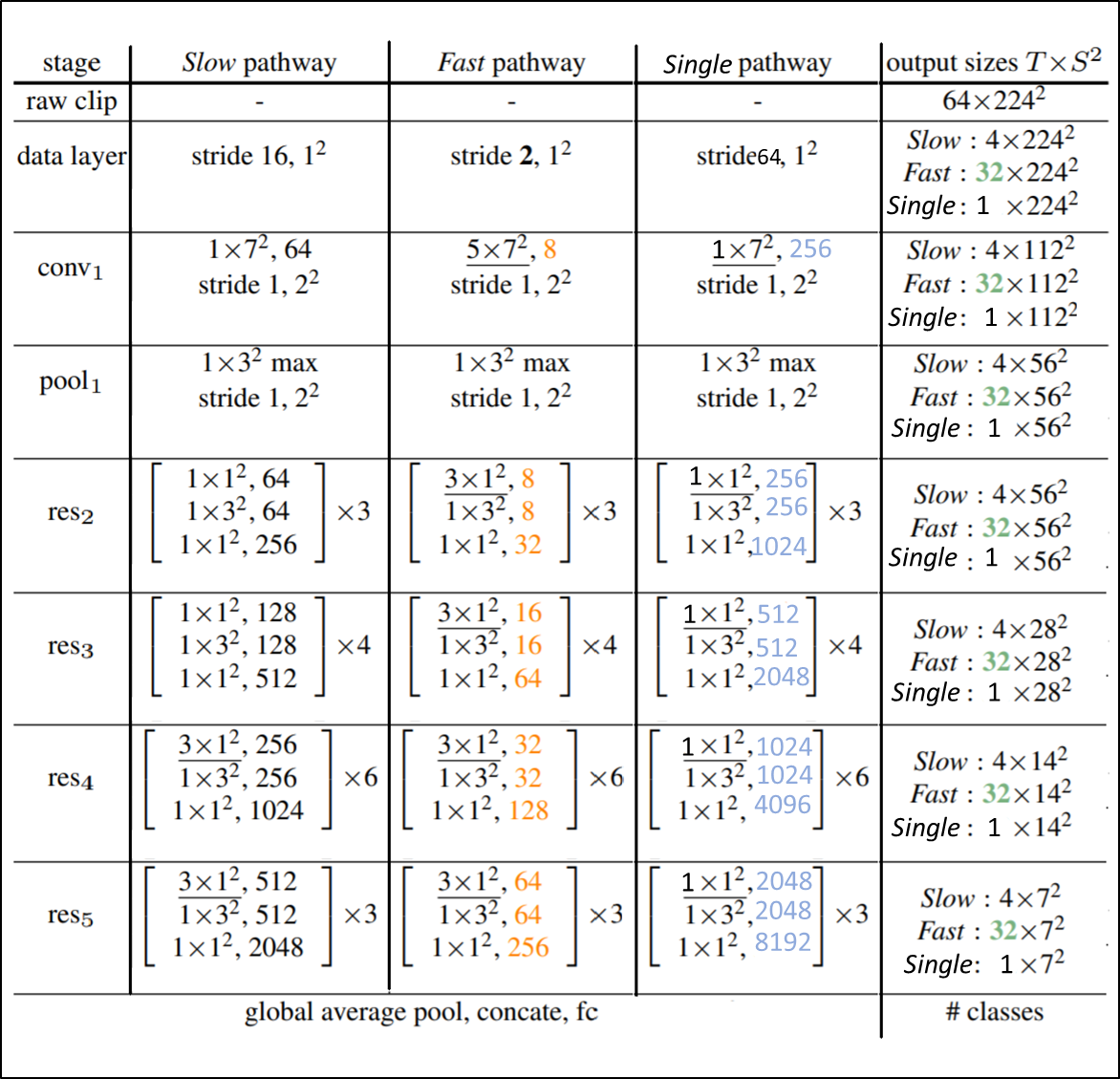}
\end{center}
  \caption{Overview of ResNet-52 for different pathways of the network encoder}
\label{fig:long}
\label{fig:onecol}
\end{figure}

\subsection{Pathway link}
Similar to \cite{2str}, \cite{yang2020temporal}, we attach one lateral connection between the three pathways for every stage. Specifically for ResNets , these connections are right after
pool1, res2, res3, and res4. The two pathways have different
temporal dimensions, so the lateral connections perform a transformation to match them. We
use unidirectional connections that fuse features of the Fast
pathway into the Slow pathway and finally Slow pathway to Single pathway. To aggregate all of features, and to ensure the compatibility of the addition between consecutive features,
during aggregation a down/up-sampling operation, we multiply by a factor.

\section{Experiments}
We evaluate our approach on four video recognition
datasets using standard evaluation protocols. For the action
classification experiments, presented in this section we consider the widely used Kinetics-600 \cite{DBLP:journals/corr/abs-1808-01340} and UCF-101 \cite{soomro2012ucf101}. For action detection experiments
in Sec. 5, we use the challenging AVA \cite{gu2018ava} dataset.
\subsection{Dataset}
In this subsection, we discuss about the different datasets we use to evaluate the performance of our networks.
\subsubsection{UCF-101}
UCF \cite{soomro2012ucf101} dataset is an open source dataset collected from videos on YouTube. It contains 101 action classes with over 13000 videos and 27 complete hours of data. The dataset consists of pragmatic videos consisting camera motion and untidy and uneven background uploaded by the user. The videos are recorded unconstrained environments and typically includes camera movement, different lighting environments, partial obstructions, low quality clips. The action categories are divided into five types - Human Object actions, Body-Motion actions, Human Human actions, Playing Musical Instruments actions, Sports based action. Categories like sports has multiple actions, where actions performed with similar background, like greenery in most cases. Some clips were captured with different illuminations, poses, and from different viewpoints. A crucial challenge of this collection of data is actions are performed in real life and are very realistic, which is significant compared to other datasets where an actor is used to perform the actions.

\begin{figure}[t]
\begin{center}
  \includegraphics[width=1.0\linewidth]{ 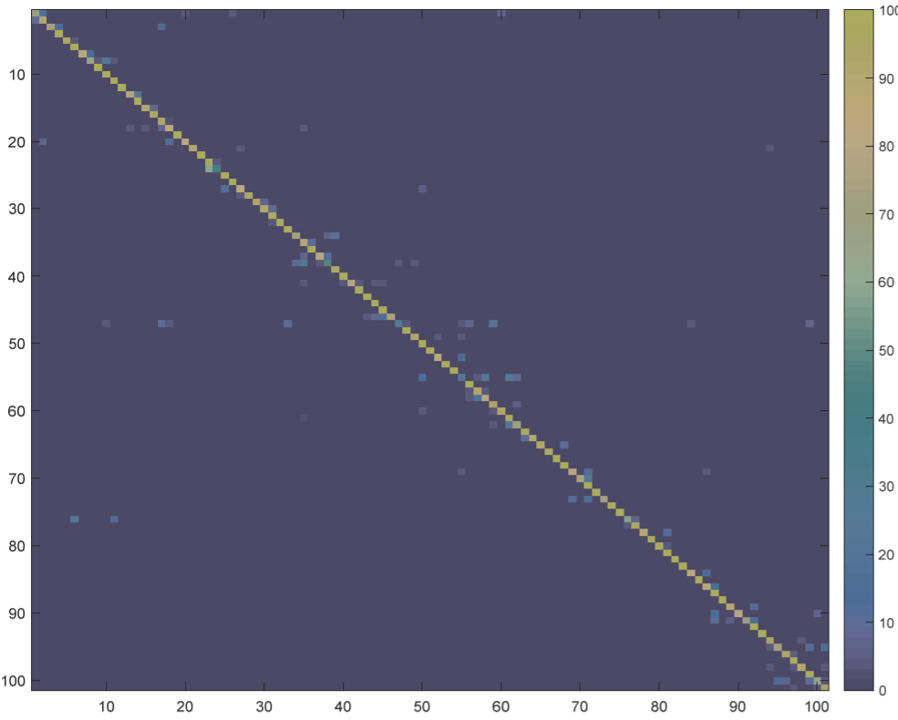}
\end{center}
  \caption{Performance of Bi-LSTM based three stream network on UCF-101 dataset, here shown as confusion matrix.}
\label{fig:long}
\label{fig:onecol}
\end{figure}

\subsubsection{Kinetics}
Kinetics is a large open source dataset by Google that around seven hundred action classes. Each class contains atleast 600 clips. Length of each video is 10s and is taken from a different YouTube video. The actions in video are human centric and cover a wide assortment of classes including human object relations such as playing instruments, human human interactions such as hugging. The videos have a variable resolution and frame rate.
\subsubsection{AVA}
The AVA dataset densely annotates 80 atomic visual actions in 430 15-minute video clips, where actions are localized in space and time, resulting in 1.58M action labels with multiple labels per person occurring frequently. The dataset is sourced from the 15th to 30th minute time intervals of 430 different movies, which given the 1 Hz sampling frequency gives us nearly 900 keyframes for each movie.
Training/validation/test sets are split at the video level, so that all segments of one video appear only in one split. There are 211k training and 57k validation video segments. The performance metric is mean Average Precision (mAP) over 80 classes, using a frame-level IoU threshold of 0.5.

\subsection{Training}

Following the setting in \cite{feichtenhofer2019slowfast}, the input frames are sampled at the $\theta_3$ and $\theta_3$ frames from the video frames. We randomly crop 224×224.
A dropout of 0.5 are adopted to
reduce overfitting. BatchNorm is not frozen.
We use a momentum of 0.9, a weight decay of 0.00001 and a
synchronized SGD training over 8 GPUs. Each GPU has
a batch-size of 8, resulting in a mini-batch of 64 in total.

\subsubsection{AVA dataset}
The AVA \cite{gu2018ava} dataset focuses on spatiotemporal localization of human actions. Hence for this dataset, unlike the other datasets, we need a detection architecture. The architecture we use is inspired from and is similar to Faster RCNN \cite{ren2015faster} and \cite{feichtenhofer2019slowfast}. We use three stream network backbones along with modified RCNNs. The fifth residual block in our implementation for AVA, we set the stride as 1 instead of 2 and dilate its filters by 2. Therefore region-of-interest (ROI) is extracted via the features of last(fifth) residual block. Like \cite{DBLP:journals/corr/abs-1808-01340}, we extend 2D ROI to 3D ROI by replicating it along temporal dimension. These features are fed through max pooling and a per-class sigmiod based classifier.

\subsection{Inference}
We uniformly sample 10 clips from a video along its temporal axis.
Following \cite{wang2018videos} and \cite{feichtenhofer2019slowfast}, we scale the shorter spatial side to 256 pixels and take 3 crops of 256×256 to cover the spatial dimensions, as an
approximation of fully-convolutional testing. We average the softmax scores for prediction.

\section{Results}
\begin{figure*}[h]
\begin{center}
  \includegraphics[width=1.0\linewidth]{ 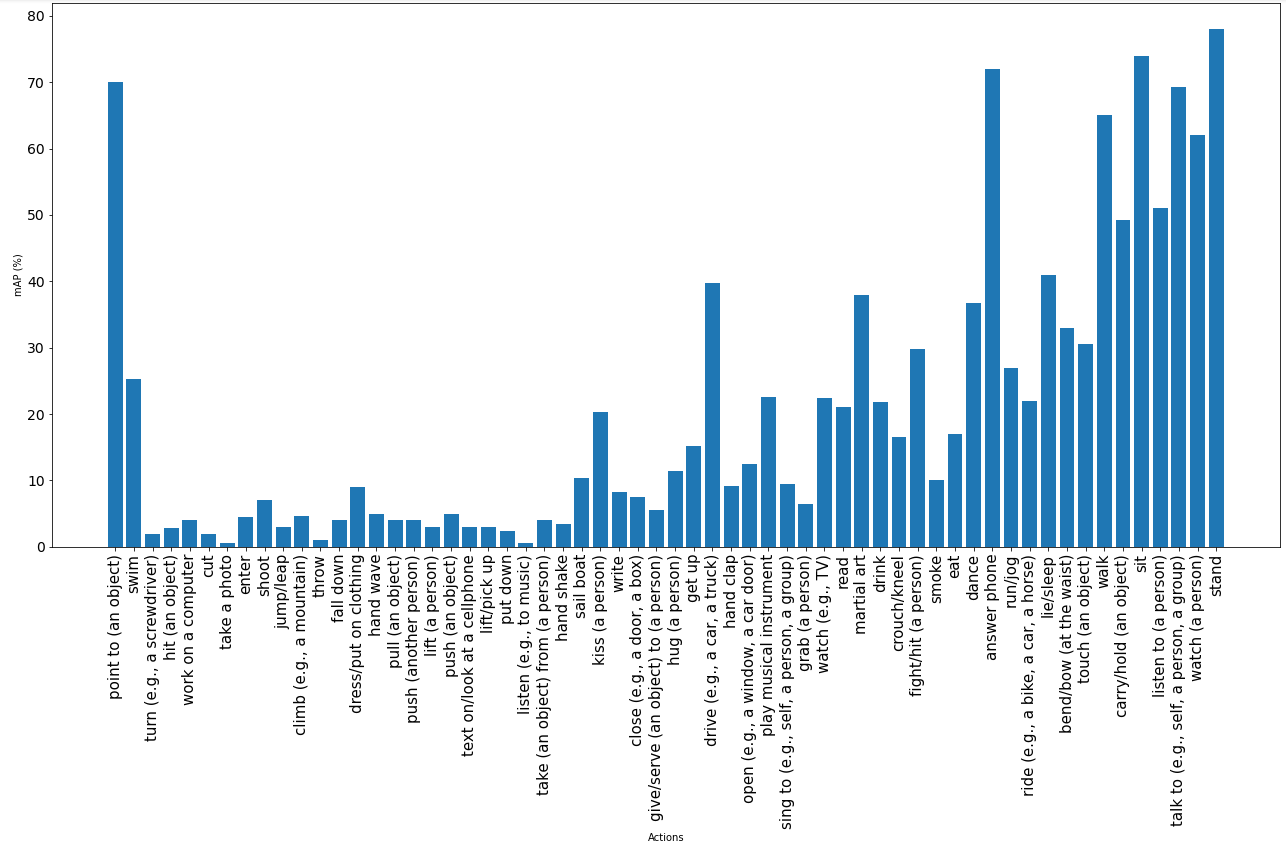}
\end{center}
   \caption{Bar chart for each action on AVA dataset, per category AP}
\label{fig:short}
\end{figure*}
We evaluate our method on baseline method and attention based model along with current state of art models on Kinetics, AVA and UCF dataset in Table 1. We observe that our model outperforms the state of art networks with attention based model performing the best. It is worth noting that we obtain this performance only using raw RGB frames as input, while prior works use RGB, flow, and in some cases audio as well.

\subsection{Baseline model for UCF-101 and Kinetics}

Table 1 shows comparison with various network architechures along with our implementation. In comparison with previous state of art our model provides 3.1$\%$ higher top-1 accuracy for Kinetics dataset and 1.4$\%$ UCF-101 dataset and. Note that our results are better even without Image-net pre training. In particular our model is 6$\%$ per better than best result of previous kind. We had experimented with ImageNet pre-training and observed that there is less than 1 percent improvement when compared to training from scratch. 

\begin{table}[]
\centering
\begin{tabular}{|l|l|l|l|}
\hline
Model                     & \multicolumn{2}{l|}{Kinetics-600} & UCF-101           \\ \hline
                          & top-1         & top-5         &          \\ \hline
Two stream \cite{2str}                & 63.2          & 79.9          & 88            \\ \hline
I3D        \cite{carreira2017quo}               & 72.1          & 89.9          & 94            \\ \hline
Two stream I3D   \cite{carreira2017quo}        & 75.7          & 90.1          & 95.6          \\ \hline
S3D          \cite{s3d}             & 69.4          & 88.0          & 95.9          \\ \hline
Nonlocal  \cite{nl}                & 77.3          & 92.6          & 93.2          \\ \hline
R(2+1)D         \cite{tran2018closer}          & 73.8          & 88.5          & 96.8          \\ \hline
STC  \cite{diba2018spatio}                     & 68.2          & 88.4          & 85.3          \\ \hline
ARTNet \cite{wang2018appearance}         & 69.2          & 88.0          & 89.7          \\ \hline
ECO    \cite{zolfaghari2018eco}                   & 70.0          & 89.4          & 90.1          \\ \hline
SlowFast       \cite{feichtenhofer2019slowfast}           & 77.0          & 92.6          & 96.8          \\ \hline
CoViAR   \cite{covair}                 & 65.4          & 75.6          & 86.4          \\ \hline
\textbf{3S+Bi-LSTM(ours)} & 80.1 & 94.0 & 98.2 \\ \hline
\textbf{3S+Attention(ours)} & \textbf{82.3} & \textbf{95.5} & \textbf{99.0} \\ \hline

\end{tabular}
\caption{A comparison of different methods incl. current state-of-art with our proposed methods on UCF-101 and Kinetics-600. Both of our methods evidently outperform the networks proposed by current literature.}
\end{table}

\subsection{Attention based model for UCF-101 and Kinetics}

As we observe in Section 6.1 that our baseline demonstrates best performance, we note that attention based three stream network outperforms that as well. This is particularly due to the ability of attention mechanism that can generate more effective temporal understandings. Our attention based model outperforms the current state of art by 5.3 $\%$. Although attention based model has enhanced temporal understandings, spatial features are not compromised as observed by lack of improvement when using ImageNet pre-trained weights. 

\subsection{Results on AVA dataset}
We compare the performance of both of our methods along with previous results. We observe that the 3 stream with Bi-LSTM has shown improvement by +0.1 mAP and our method with attention mechanism shows improvement by +6.1 mAP. It is important to note that the methods with flow stream as well can double computational cost. We observe that Kinetics pre-training shows a great improvement in accuracy too with +6.8 mAP. 
\begin{table}[h]
\centering
\begin{tabular}{|l|l|}
\hline
\textit{Model(+our version RCNN)} & \textit{AVA (mAP \%)} \\ \hline
Two stream   \cite{2str}                     & 7.4                   \\ \hline
I3D              \cite{carreira2017quo}                 & 14.5                  \\ \hline
Two stream I3D    \cite{carreira2017quo}                & 15.8                  \\ \hline
S3D           \cite{s3d}                     & 17.2                  \\ \hline
Nonlocal  \cite{nl}                         & 20.0                  \\ \hline
R(2+1)D    \cite{tran2018closer}                        & 21.1                  \\ \hline
STC       \cite{diba2018spatio}                        & 15.6                  \\ \hline
ARTNet   \cite{wang2018appearance}                          & 13.2                  \\ \hline
ECO    \cite{zolfaghari2018eco}                           & 12.8                  \\ \hline
SlowFast   \cite{feichtenhofer2019slowfast}                       & 26.3                  \\ \hline
CoViAR       \cite{covair}                     & 21.9                  \\ \hline
\textbf{3S+bi-LSTM(ours)}         & 26.4                 \\ \hline
\textbf{3S+Attention(ours)}       & \textbf{32.2}         \\ \hline
\end{tabular}
\caption{A comparison of different methods incl. current state-of-art with our proposed methods on AVA. While our baseline model shows very small improvement over AVA dataset, the attention head shows significant improvement from the current state-of-art.}
\end{table}

\section{Ablation Study}
All of our models have used class agnostic regression data augmentation (and Kinetics pre-training for AVA models) techniques we observed early on to be critical for good performance. 

A key intuition for designing the different pathways was to use lower channel capacity for Fast pathway. We experimented with different values of $\theta_2$ ranging from 4 to 32. As expected the accuracy (measured on Kinetics) decreased with the values of $\theta_2$.  But this also increased the GFLOPs. Hence to balance, we choose 16 frames.
\begin{table}[h]
\centering
\begin{tabular}{|l|l|l|}
\hline
\textit{$\theta_2$ value} & \textit{top-1 ($\%$)} & \textit{top-5 ($\%$)} \\ \hline
4                   & 79.9           & 94.8           \\ \hline
6                & 80.2           & 94.9          \\ \hline
12                  & \textbf{80.3}           &  \textbf{95.6}          \\ \hline
16                 & 80.0           & 95.3           \\ \hline
32                  & 80.1           & 95.0           \\ \hline

\end{tabular}
\caption{Different methods to laterally fuse}
\end{table}

We also compared the effectiveness of adding temporal block i.e. Bi-LSTM and attention blocks to our three stream network on the AVA dataset. Temporal blocks improves performance on 74 out of 80 actions as compared to vanilla three stream networks. The categories that showed major gains were: \textit{run/jog} (21.2 AP), \textit{hand wave} (15.7 AP), \textit{hand clap} (24.6 AP), 
\textit{eat} (13.5 AP). These are categories where modeling
dynamics are of vital importance.

Our three stream CNN encoder can have any architecture type. We experimented with various networks. We observe that Res-152 performs best as the backbone as compared to InceptionV3\cite{szegedy2016rethinking}, Res-101 \cite{resnet}, Res-52, EfficientNet B5 \cite{tan2019efficientnet}, and EfficientNet B7. But we select ResNet-52 as trade-off against the GFLOPs trade-off.

\begin{figure}[h]
\begin{center}
  \includegraphics[width=1.0\linewidth]{ 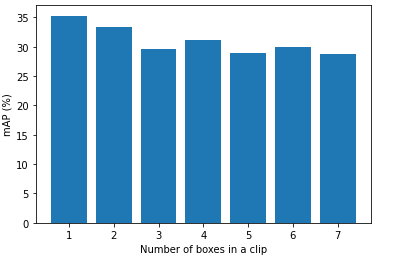}
\end{center}
  \caption{Performance of Attention based three stream network with respective to the number of bounding boxes in the clip. }
\label{fig:long}
\label{fig:onecol}
\end{figure}

\begin{figure}[h]
\begin{center}
  \includegraphics[width=1.0\linewidth]{ 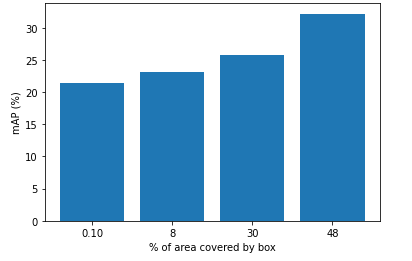}
\end{center}
  \caption{Performance of Attention based three stream network with respective to the area covered by bounding boxes of the clip. }
\label{fig:long}
\label{fig:onecol}
\end{figure}

\section{Conclusion}
In this paper we proposed two network architechures that signifies improvement in spatio-temporal understanding of actions in a video.  It achieves state-of-the-art accuracy for video action classification and detection. We hope three stream network will foster further research in action and video understanding.


\section{Acknowledgements}
We would like to thank Dr. Carlo Ciliberto for his very helpful insights and Imperial College London for hardware support in initial project phase.

{\small
\bibliographystyle{ieee_fullname}
\bibliography{cvpr}
}

\end{document}